\documentclass[PCJ,Unicode,screen]{cedram}

\usepackage[natbib=true,mincitenames=1,maxcitenames=2,maxbibnames=20,minbibnames=20,uniquelist=false,uniquename=false,style=pcj]{biblatex}
\usepackage{lipsum}
\usepackage{setspace}

\usepackage{caption}
\usepackage{subcaption} 
\usepackage{amsfonts}
\usepackage{wasysym}
\usepackage{textcomp}
\usepackage{algorithm}
\usepackage{algpseudocode}
\usepackage{amssymb}

\providecommand*{\noopsort}[1]{}

\PCIcorresp{Valentin.Guien@uca.fr}

\title[Sample for the template]{Detection of anomalies in cow activity using wavelet transform based features}
\author{\firstname{Valentin} \lastname{Guien}}
\address{Université Clermont Auvergne, CNRS, Clermont Auvergne INP, Mines Saint-Étienne, LIMOS, 63000
	Clermont-Ferrand, France}
\email[V. Guien]{Valentin.Guien@uca.fr}
\author{\firstname{Violaine}  \lastname{Antoine}}
\address{Université Clermont Auvergne, CNRS, Clermont Auvergne INP, Mines Saint-Étienne, LIMOS, 63000
	Clermont-Ferrand, France}
\email[V. Antoine]{Violaine.Antoine@uca.fr}
\author{\firstname{Romain} \lastname{Lardy}}
\address{Université Clermont Auvergne, INRAE, VetAgro Sup, UMR Herbivores, F-63122 Saint-Genès-Champanelle, France}
\email[R. Lardy]{Romain.Lardy@inrae.fr}
\author{\firstname{Isabelle} \lastname{Veissier}}
\address{Université Clermont Auvergne, INRAE, VetAgro Sup, UMR Herbivores, F-63122 Saint-Genès-Champanelle, France}
\email[I. Veissier]{Isabelle.Veissier@inrae.fr}

\author{\firstname{Luis E} \lastname{C Rocha}}
\address{Ghent University, Department of Economics, 9000 Ghent,
	Belgium}
\address{Ghent University, Department of Physics and Astronomy, 9000 Ghent, Belgium}
\email[L. Rocha]{luis.rocha@ugent.be}

\begin{abstract}
	
	In Precision Livestock Farming, detecting deviations from optimal or baseline values – i.e. anomalies in time series – is essential to allow undertaking corrective actions rapidly. Here we aim at detecting anomalies in 24 h time series of cow activity, with a view to detect cases of disease or oestrus. Deviations must be distinguished from noise which can be very high in case of biological data. It is also important to detect the anomaly early, e.g. before a farmer would notice it visually. Here, we investigate the benefit of using wavelet transforms to denoise data and we assess the performance of an anomaly detection algorithm considering the timing of the detection.
	We developed features based on the comparisons between the wavelet transforms of the mean of the time series and the wavelet transforms of individual time series instances. We hypothesized that these features contribute to the detection of anomalies in periodic time series using a feature-based algorithm. We tested this hypothesis with two datasets representing cow activity, which typically follows a daily pattern but can deviate due to specific physiological or pathological conditions. We applied features derived from wavelet transform as well as statistical features in an Isolation Forest algorithm. We measured the distance of detection between the days annotated abnormal by animal caretakers days and the days predicted abnormal by the algorithm. The results show that wavelet-based features are among the features most contributing to anomaly detection. They also show that detections are close to the annotated days, and often precede it.
	In conclusion, using wavelet transforms on time series of cow activity data helps to detect anomalies related to specific cow states. The detection is often obtained on days that precede the day annotated by caretakers, which offer possibility to take corrective actions at an early stage.

\end{abstract}
\begin{document}
	
	\maketitle
	\nolinenumbers
	
	\section{Introduction}

	Precision Livestock Farming is based on monitoring animals or their environment thanks to sensors to support farmers’ decisions (\cite{Berckmans2014PrecisionLF}). It requires that deviations from the optimal or baseline values – i.e.,  anomalies in time series – are detected. There is evidence that the behavior of animals is modified under certain conditions such as oestrus, parturition, disturbances or diseases. The variations in an animal activity during 24 h seem very sensitive to the condition of that animal (\cite{wagner_detection_2021}). Still, it remains challenging to detect anomalies from the 24-hour variations of activity with high sensitivity and accuracy (e.g. \cite{lardy_discriminating_2023}). This is likely due to the high variability between and within individuals (\cite{Fuchs2022}), which makes it essential to denoise the data. 
	
	Anomaly detection has been extensively studied (\cite{chandola_anomaly_2009}) across various fields such as electricity consumption (\cite{himeur_artificial_2021}), card fraud (\cite{hilal_financial_2022}), healthcare (\cite{faverjon_choosing_2018,paragliola_gait_2018}), and agriculture (\cite{wagner_detection_2021,CATALANO2022103750}). In some domains, prior knowledge provides insights into the characteristics of the time series, allowing the choice of specific algorithms for anomaly detection. In this paper, we focus on the time series of cows’ activity with the following characteristics: the activity follows a regular periodicity but is also highly variable; thus, distinguishing anomalies from spontaneous variations – i.e. noises – becomes challenging. Providing insight about how anomalies manifest in a time series – to explain the decision of the model and to understand the onset of the anomaly – can be done by transforming the time series into a set of features and using these features as inputs in a standard anomaly detection algorithm (\cite{jiang_gan-based_2019}). 
	
	For instance, statistical features can be computed from time series, and a Principal Component Analysis (PCA) can be performed on these features (\cite{7395871}). The study from \cite{7395871} detected abnormal time series using the first two principal components. PCA speeds up computation but comes at the cost of features interpretability. Random Forest has been applied to predict different conditions of cows from their daily activity (Lardy et al., 2023). This previous study extracted statistical and Fourier transform features from each 24-hour time window. The features described intensity (e.g., mean, maximum and mode, quantiles), variability (e.g., standard deviation, root mean square of differences between successive hours) and periodicity (autocorrelations, Fourier transform). The contribution of each feature was registered as weight, allowing us to identify which features contribute the most to the classification. The use of the Fourier transform may be outperformed by the Discrete Wavelet Transforms (DWT) that allow not only a frequency analysis of the signal but also a time-localization of the changes in the time series (\cite{57199,1992AnRFMF}). DWT is commonly used for data compression (\cite{chiarot_time_2023}), and by extension, for denoising, which makes these methods promising for processing data on animal behavior.
	
	Wavelet transforms can be incorporated at various stages of an anomaly detection process. For example, the Wavelet-based Anomaly Detector (WAD) (\cite{zhang_detection_nodate}) applies wavelet transform to time series data to obtain the Trend data and the Residual data. Anomalies are then identified in the Residual data from days when a value exceeds a threshold determined by the statistical distribution of the Residual data. This approach shows the ability of wavelets to capture explicit trends in data. However, WAD primarily focuses on detecting anomalies as isolated points, whereas, in many time series applications, anomalies are represented as sequences of points rather than single points, as in the case of the daily fluctuations of the activity of animals. The AutoWave method (\cite{yao_regularizing_2023}) is a regularizing auto-encoder that uses the DWT in its frequency regularizer to reduce the latent space, so the anomalies are poorly reconstructed. This method highlights the strength of the wavelet transform in detecting specific patterns in time series. However, while AutoWave learns features during the training, its input consists of a sequence of values, limiting the interpretability of the detection due to the lack of explicit features. 
	
	In Precision Livestock Farming, the detection must be sensitive (it captures most of the anomalies) and specific (it does not produce false alarms) and occurs early so that farmers can make management decisions quickly. An animal's behaviour can change several days before clinical signs of disease are observed (\cite{MARCHESINI201841, WOTTLIN2021104488}) or one day before ovulation in case of oestrus. In this paper, we evaluate the effectiveness of DWT-based features in reducing the inherent variability of the time series of cows’ activity while retaining the essential information needed to detect anomalies (here, diseases and oestrus ). We also investigate ways to measure how early detection takes place by measuring the delay between the detection and the true anomaly (here, that is the time when caretakers detect a disease or oestrus by observing animals).

	\section{Materials and Methods}
	
	\subsection{Time series preprocessing and feature generation}

	\subsubsection{Periodic time series with labels}
We consider a data set $D$ composed of time series data for $n_I$ individuals. Each individual has their own time series, such that $D=[X_1,\dots,X_{n_I}]$ where $X_j = [x_{1,j},\dots,x_{n_j,j}] \in \mathbb{R}^{n_j}$ is the time series of individual $j$ ($1 \le j \le n_I$) consisting of $n_j$ values. The values of $n_j$ can vary because the number of observations differs between individuals. We also assume these time series are periodic with a period length $\Delta T$, i.e. $[x_{i,j},\dots,x_{i+p-1,j}] \allowbreak \HF \allowbreak[x_{i+\Delta T,j},\dots,x_{i+2\Delta T-1,j}]$. 

Each value $x_{i,j}$ of the time series is associated with a label $y_{i,j} \in \{N,A,F\}$. If $y_{i,j} = N$ (or $y_{i,j} = A$), then $x_{i,j}$ is labelled as \textit{normal} (or \textit{abnormal}). If $y_{i,j} = F$, the value $x_{i,j}$ is labelled as \textit{fuzzy}, indicating uncertainty regarding the true label. Other labels, i.e. \textit{normal} and \textit{abnormal}, typically correspond to consecutive values, as they represent states that persist longer than the interval between points in the time series.
	
	
	\subsubsection{Windows}
	
\begin{figure}[ht!]
		\includegraphics[width=\textwidth]{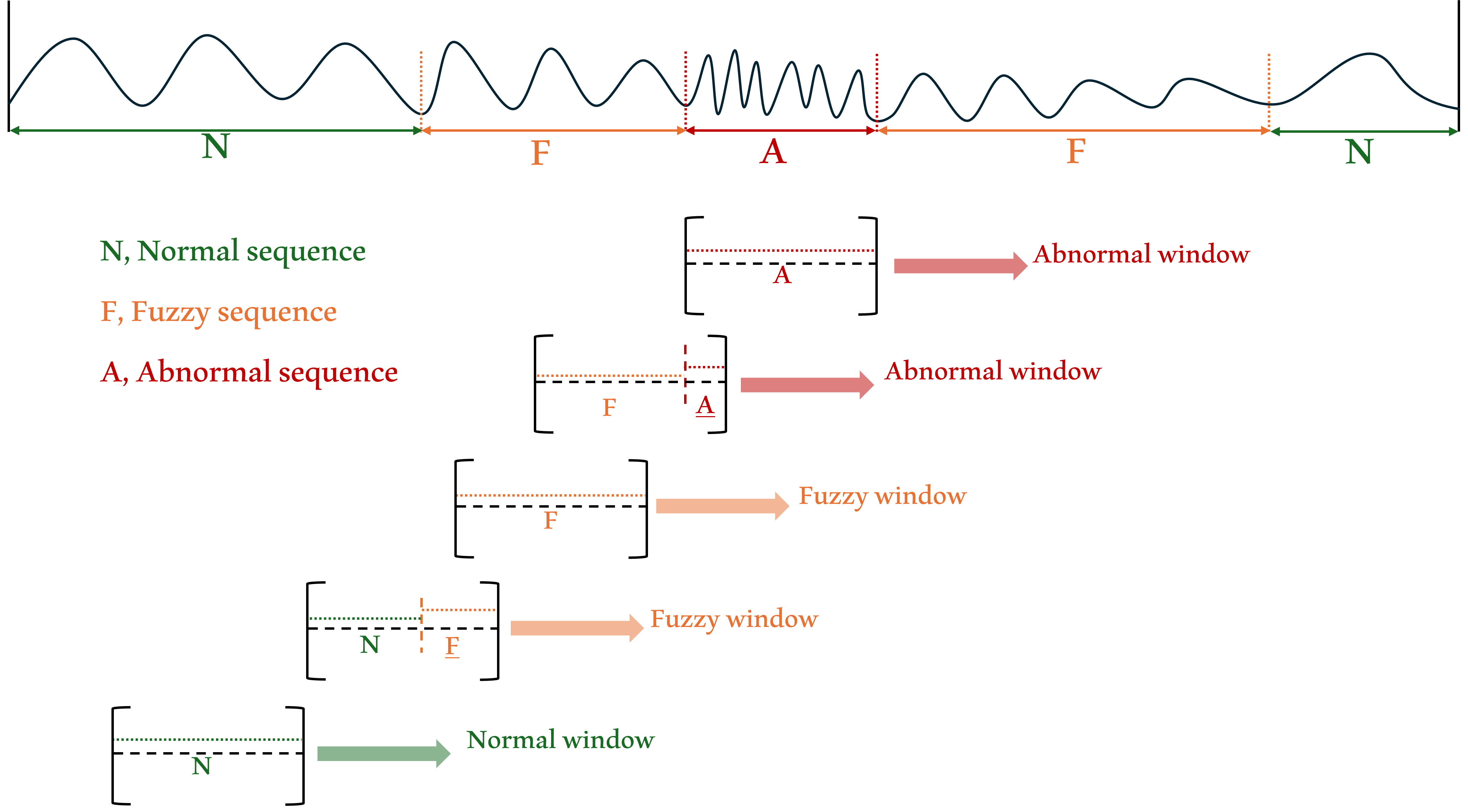}
		\caption{Window labelling. If at least one data from an abnormal period is in the window, the window is labelled abnormal. If at least one data from a fuzzy period is in the window, without any abnormal data, the window is labelled fuzzy. Otherwise, the window is labelled normal.} \label{windows}
	\end{figure}

	To follow the evolution in the time series, we consider windows of size $q$ through the whole time series. 
	Each window $W_{i,j} \in \mathbb{R}^q$ constitute a set of values such that $W_{i,j} = [x_{i,j},...,x_{i+q-1,j}]$. 
	By extension to labels on values, we define for each window $W_{i,j}$ a label $Y_{i,j} \in \{\textrm{\textit{Normal}},\textrm{\textit{Abnormal}},\textrm{\textit{Fuzzy}}\}$.
	The window is labelled depending on whether the window contains at least one abnormal or fuzzy values. 
	If $\exists k \in  [\![i,i+q-1]\!]$ such that $y_{k,j} = A$, then $Y_{i,j} = \textrm{\textit{Abnormal}}$. 
	If $\forall k \in  [\![i,i+q-1]\!], y_{k,j} \neq A$, and $\exists k \in  [\![i,i+q-1]\!]$ such that $y_{k,j} = F$, then $Y_{i,j} = \textrm{\textit{Fuzzy}}$. 
	Else if $\forall k \in  [\![i,i+q-1]\!], y_{k,j} = N$ then $Y_{i,j} = \textrm{\textit{Normal}}$. 
	The windows are labelled considering the `worst-case scenario'. 
	This determination of the windows labels is illustrated in Figure \ref{windows}.
	
	We define two windows $W_{i_1,j_1},W_{i_2,j_2} \in \mathbb{R}^q$ as consecutive if $i_1 = i_2-1$ and $j_1 = j_2$. A series of consecutive windows corresponds to a set of windows $[W_{i_1,j},\dots,W_{i_n,j}]$ where $\forall k \in [\![1,n]\!], W_{i_k,j}$ and $W_{i_{k+1},j}$ are consecutive. As the windows slide through the time series, all consecutive windows are identified. Series of consecutive windows are formed to ensure that they remain contiguous when selected for the training or testing sets.

	\subsubsection{Construction of the training and testing sets}
	
	The anomaly detection algorithm is trained on data that does not contain anomalies, allowing it to learn the normal behavior of the time series. When data in the testing set significantly deviate from this learned pattern, the algorithm flags it as an anomaly. Both training and testing sets are built using windows of the time series, considering not just the labels but also the sequential nature of the windows.
	
	Let $\mathcal{W}_{Abnormal}$, $\mathcal{W}_{Normal}$ and $\mathcal{W}_{Fuzzy}$ be the sets of windows labelled as Abnormal, Normal and Fuzzy, respectively. Let $\mathcal{W}_{train}$ and $\mathcal{W}_{test}$ represent the sets of windows used for building the training set and the testing set, respectively. All windows in $\mathcal{W}_{Abnormal}$ and $\mathcal{W}_{Fuzzy}$ belong to $\mathcal{W}_{test}$. 
	
	Since we created instances using a sliding window with a small sliding step size, the cases with only a few steps of difference are naturally similar. To avoid biases when evaluating the quality of anomaly detection, such instances, referred to as consecutive windows, are kept together either in the training or testing sets.
	
	
	The set $\mathcal{W}_{Normal}$ is composed of sets of  consecutive windows $\mathcal{SW}_{Normal} = [SW_1,\dots,SW_l]$, where $SW_k$ represents a set of consecutive windows labeled as \textit{Normal}. The testing set $\mathcal{W}_{test}$ is then completed using consecutive windows to create a testing set with a balanced number of anomaly and normal instances. Since the size of the sets $SW_k$ and the size of  $\mathcal{W}_{Abnormal}$ depend on the datasets, we approach the balance with the following procedure. While there are more abnormal windows than normal windows in $\mathcal{W}_{test}$, we randomly select a set of consecutive windows from $\mathcal{SW}_{Normal}$ and add all the windows it contains in $\mathcal{W}_{test}$. The overall procedure for constructing the training and testing sets is described in Algorithm \ref{alg}.
	
	
	\begin{algorithm}[ht!]
		\caption{Construction of training set and testing sets}
		\label{alg}
		\begin{algorithmic}[1]
			\Function{BuildTrainTest}{$\mathcal{W}_{Normal}, \mathcal{W}_{Abnormal}, \mathcal{W}_{Fuzzy}$}
			\State Initialize $\mathcal{W}_{train} \gets \emptyset$
			\State Initialize $\mathcal{W}_{test} \gets \mathcal{W}_{Abnormal} \cup \mathcal{W}_{Fuzzy}$
			\State Get $\mathcal{SW}_{Normal}$ from $\mathcal{W}_{Normal}$
			\While{$|\mathcal{W}_{test} \cap \mathcal{W}_{Normal}| < |\mathcal{W}_{Abnormal}|$}
			\State Select randomly $SW$ from $\mathcal{SW}_{Normal}$
			\For{$W \in SW$}
			\State $\mathcal{W}_{test} \gets \mathcal{W}_{test} \cup \{W\}$
			\EndFor 
			\State $\mathcal{SW}_{Normal} =  \mathcal{SW}_{Normal} \smallsetminus \{SW\}$
			\EndWhile
			\State ${W}_{train} \gets \mathcal{W}_{Normal} \smallsetminus (\mathcal{W}_{test} \cap \mathcal{W}_{Normal})$
			\State \Return ${W}_{train},{W}_{test}$
			\EndFunction
		\end{algorithmic}
		
	\end{algorithm}

	The number of fuzzy windows is not considered when constructing the training and testing sets, as fuzzy windows represent sequences where information about the anomalies is uncertain.
	
	\subsubsection{Wavelet based features}

\begin{figure}[ht!]
		\centering
		\includegraphics[width=0.9\textwidth]{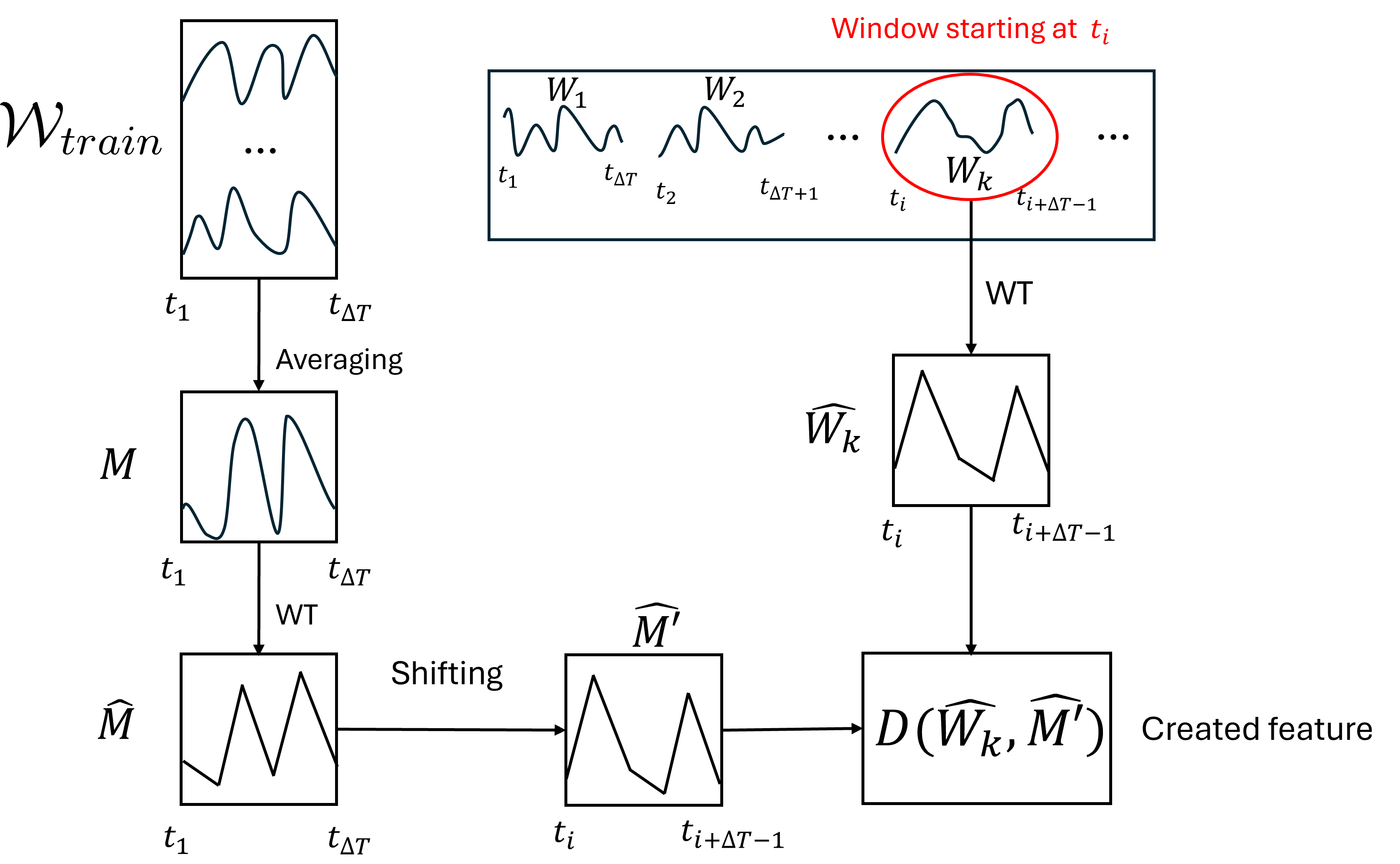}
		\caption{Pipeline for the creation of wavelet-based features.} \label{features}
	\end{figure}

	The wavelet transform is appropriate for analyzing periodic time series with high variability. Let the window size $q=\Delta T$ be chosen to correspond to the length of the period, and let $\{t_1,\dots,t_{\Delta T}\}$ represent the timestamps where each window can start. These timestamps can shift, meaning that a window starting at $t_i$ corresponds to measurements taken at $\{t_i,\dots,t_{\Delta T},t_{1},\dots,t_{i-1}\}$. Figure \ref{features} illustrates the pipeline for creating wavelet-based features.
	
	Let us consider $M = [m_{t_1},\dots,m_{t_{\Delta T}}]$ the average period of the normal time series from the training set $\mathcal{W}_{train}$. Each point $m_{t_k}$ is the mean of the values taken at $t_k$ in the selected windows where the timestamp corresponds to $t_1$.
	
	Next, the DWT is applied to $M$, yielding the approximated signal $\hat{M} = [\widehat{m}_{t_1},\allowbreak\dots,\allowbreak\widehat{m}_{\Delta T}]$. This signal represents the transformed average period, serving as a reference to generate the wavelet-based features.
	
	Let $W_{k}$ be a window containing a sequence of values, and let $\widehat{W_k}$ denote the DWT applied on $W_{k}$. The objective is to compare the wavelet transform of this window $\widehat{W_k}$ with the wavelet transform of the average period. Since $W_{k}$ starts at timestamp $t_i$, $\widehat{M}$ is shifted to align with $t_i$, resulting in a new sequence $\widehat{M'}$.
	
	 $\widehat{M'}$ and $\widehat{W_k}$ are compared using the Euclidean distance between the two windows. A large (vs low) distance indicates significant (vs no significant) deviation from the average normal period.

	The DWT of a signal requires two parameters. The first is the choice of the wavelet used for the transformation. Different types of wavelet can be grouped into families (\cite{388960}). The second parameter is the level of approximation of the signal. In a DWT, the signal passes through a series of low- and high-pass filters (\cite{Shensa1992TheDWT}). The level of approximation corresponds to how many times the signal undergoes filtering. The maximum approximation level depends on the chosen wavelet and the length of the signal.

	\subsection{Application to disease and oestrus detection from cow behavior}
	
	\subsubsection{Activity level data}
	
	The data used for the study are time series of dairy cow behavior (\cite{lardy_understanding_2022}). Two data sets from experimental farms are used (Table \ref{datasets}). The data concerns cows in a barn, with a sensor on their collar tracing their activity. The sensor locates the cow in real-time. Three areas are considered in the barn: the feeding table, the cubicles, and the alleys. The current activity of the cow is estimated from its position. If the cow is near the feeding table, we assume the cow eats. If the cow is in a cubicle, we assume the cow is resting. If the cow is in the alley, we assume the cow is walking or standing still. A Factorial Analysis of Correspondence is performed on the three activities to get their weighted contribution for the hourly activity (\cite{VEISSIER20173969}). For each cow $i$ and hour $h$ of the day, an Activity Level ($AL_i(h)$) is thus computed from the duration of activities within that hour:
	
	\begin{equation}
		AL_i(h) = -0.23\times(\textrm{time spent resting}) + 0.16\times(\textrm{time spent in alleys}) + 0.42\times(\textrm{time spent eating})
	\end{equation}

\begin{figure}[ht!]
		\centering
		\includegraphics[width=1.0\textwidth]{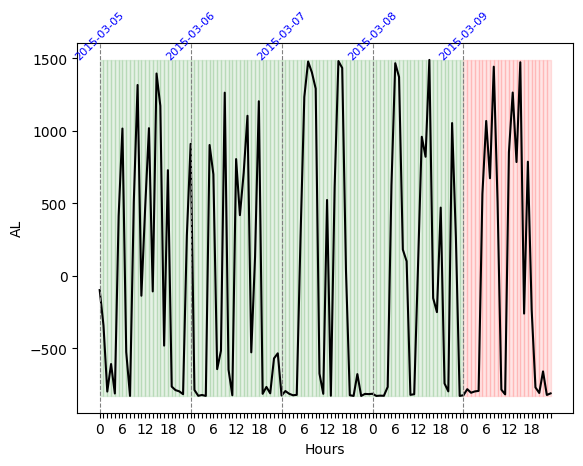}
		\caption{Activity Level (AL) of Cow n°7163 from Dataset 2 from March 5, 2013 to March 9, 2013. The green background corresponds to days with no recorded anomaly, and the red background corresponds to the day with an anomaly recorded by caretakers.} \label{TS_cows}
	\end{figure}

	The data is recorded as a time series representing the values of $AL(h)$, where $h$ is the hour of the day. $AL_i(h)$ depicts the evolution of the cow's level of arousal (\cite{VEISSIER20173969}). The time series are periodic, and the length of the period is $\Delta T=24$ (i.e. one day). Specific states of the cow (e.g. disease, oestrus, calving) and events (e.g. mixing, management change, disturbances due to external events such as electric failures) occurring in the barn are recorded manually by caretakers. We expect these events to change the cows' activity, resulting in the behavioral anomaly we want to detect in the time series using our proposed methodology (\cite{VEISSIER20173969}). Figure \ref{TS_cows} shows a portion of a time series without anomaly of a certain cow.
	These time series data contain much noise because the activity can greatly vary between cows and also within cows between days. In addition, the exact onset of the anomaly is unknown: the caretakers can miss an anomaly or the behavior of a cow can start changing before it becomes visible by caretakers.

	\begin{table}[ht!]
		\caption{Summary information of the datasets.}
		\label{datasets}
		\centering
		\begin{tabular}{|l|l|l|}
			\hline
			& Dataset 1 & Dataset 2 \\
			\hline
			Source & INRAE Herbip\^ole & INRAE Herbip\^ole\\
			Experimentation length & 2 months & 6 months\\
			Start date & 2015-03-02&25-10-2018 \\
			End date & 2015-04-30&17-04-2019 \\
			Number of cows & 28 & 28\\
			Number of cow $\times$ hour & 40 246& 107 665 \\
			Rate of missing data & 0.2\% & 12.4\%\\
			Rate of specific states &  0.7\%& 2.2\%\\
			\hline
		\end{tabular}
	\end{table}
	
	Days with events are removed from the datasets because the aim is to detect cow states. Any day $\times$ cow annotated with a specific state is considered abnormal. Days $\times$ cows with an average activity level superior to 1000 are also removed because they correspond to a non-functioning sensor. When a cow is considered abnormal for a specific day, we assume $\Delta \tau_b$ days before and $\Delta \tau_a$ days after the abnormal day as uncertain days where the cow's behavior can be modified (\cite{wagner_detection_2021}). These days are annotated as fuzzy days. The number of fuzzy days preceding or following an abnormal day depends on the specific state. In the case of an oestrus, we consider $\Delta \tau_b = 2$ and $\Delta \tau_a = 1$. In the case of calving, mastitis, lameness and other diseases, we consider $\Delta \tau_b = 2$ and $\Delta \tau_a = 7$. In the case of accidents and LPS injection, we consider $\Delta \tau_b = 0$ and $\Delta \tau_a = 7$. Finally in the case of acidosis, we consider $\Delta \tau_a = 0$ and $\Delta \tau_b = 1$.

	\subsubsection{Features to describe time series}
	
	We tested five families of wavelets: Haar, Daubechies, Coiflet, Biorthogonal, and Reverse Biorthogonal.
	
The Haar wavelet is the oldest and simplest, defined as a step function. Its simplicity makes it an ideal starting point for studying wavelet transforms. Due to its structure, the Haar wavelet allows for various levels of approximation, offering a wide range of approximations for time series data. It is commonly used for denoising (\cite{5193046,8397456}) and for performing similarity searches in time series (\cite{popivanov_similarity_2002}).	

The Daubechies wavelets (\cite{vonesch_generalized_2007}) form a family of wavelets characterized by specific mathematical properties related to vanishing moments. These wavelets have been widely applied in various studies, including disease detection, such as Parkinson's disease (\cite{8930802,Zayrit2020DaubechiesWC}).

The Coiflet wavelets have been used for anomaly detection in network traffic (\cite{lu_network_2008}), for which they have been proven to be among the most effective wavelets. Coiflet wavelets are also used for denoising (\cite{tikkanen_nonlinear_1999}).

Biorthogonal and Reverse Biorthogonal wavelets are less common in the literature on wavelet transform applications. However, \cite{guo_biorthogonal_2019} showed the advantages of Biorthogonal wavelets when dealing with imperfect reconstructions after compression and noise filtering in 3D models. Biorthogonal wavelets are also used in similarity searches in time series (\cite{popivanov_similarity_2002}).
	
For each wavelet family considered, we selected the wavelets and approximation levels so that the DWT is feasible on a window of size 24 with the specified parameters (e.g., the Haar wavelet is used in four features, and the approximation level used ranges from 1 to 4). In total, 23 wavelet-based features are generated (Table \ref{wavelets_feat}).

	
	\begin{table}[ht!]
		\caption{Wavelet-based features.}
		\centering
		\begin{tabular}{|l|p{40mm}|}
			\hline
			Wavelet name (acronym) &  Levels of approximation used for this paper\\
			\hline
			Haar & 1,2,3,4 \\
			Daubechies 2 (db2) &1,2,3 \\
			Daubechies 3 (db3) & 1,2 \\
			Coiflet 1 (coif1)  & 1,2\\
			Biorthogonal wavelet 1.3 (bior1.3) & 1,2 \\
			Biorthogonal wavelet 2.2 (bior2.2)& 1,2 \\
			Biorthogonal wavelet 3.1 (bior3.1)& 1,2,3 \\
			Reverse biorthogonal wavelet 2.2 (rbio2.2)& 1,2\\
			Reverse biorthogonal wavelet 3.1 (rbio3.1)& 1,2,3\\
			\hline
		\end{tabular}
		\label{wavelets_feat}
	\end{table}

	\subsubsection{Other features}
	The wavelet-based features are complemented by 27 features derived from descriptive statistics (Table \ref{stat_features}). We use all the features from \cite{lardy_discriminating_2023}, excluding the harmonics from the Fourier transform due to redundancy with the wavelet transforms.
	
	\begin{table}[ht!]
		\caption{Features derived from descriptive statistics.}
		\centering
		\begin{tabular}{p{72mm}|l}
			\textbf{Definition} & \textbf{Name} \\ \hline
			Minimum activity over the 24h & Minimum\\ \hline
			Maximum activity over the 24 h & Maximum\\ \hline
			Mean activity over the 24h & Mean\\ \hline
			Square root of the mean of squared activities over the 24h & RMS\\ \hline
			Standard deviation of the activity over the 24h & STD\\ \hline
			Mean of the standard deviation of the 4 non-overlapping 6h windows composing the 24h & MeanSTD6h \\ \hline
			Standard deviation of the mean of the 4 non-overlapping 6h windows composing the 24h & STDMean6h\\ \hline
			Root mean square of successive differences, i.e. the differences between the activity at a given hour $h$ and the activity at the next hour $h + 1$ & RMSSD \\ \hline
			Most common value over the 24h & Mode\\ \hline
			10\% and 90\% quantiles, calculated from the values that divide the hours into 10 equal groups from lower to higher activity. Q10, maximum values of the lower group; Q90, maximum value of the last but one higher group & Q10, Q90 \\ \hline
			25\%, 50\% and 75\% quantiles, calculated from the values which divide the hours into 4 equal groups from lower to higher activity. Q25, maximum values of the lower group; Q50, median; Q75, maximum value of the last but one higher group & Q25, Median, 75\\ \hline
			Symmetry of the activity distribution over the 24h & Skewness \\ \hline
			Tailedness of the activity distribution over the 24h & Kurtosis \\ \hline
			Autocorrelation, i.e. the correlation between the activity at any hour $h$ and the activity at hour $h + d$, where $d$ is a fixed interval (1h, 2h, ... 11h) & Autocorr1-11 \\ \hline
			
		\end{tabular}
		\label{stat_features}
	\end{table}
	
	
	\subsubsection{Anomaly detection algorithm}
	
	The anomaly detection algorithm used is Isolation Forest (\cite{liu_isolation_2008}). The algorithm is based on the construction of Isolation Trees. For a given tree, each node represents a split on random value of a random feature. Each window of 24 h is isolated in a root of the tree. The Isolation Forest builds a defined number of Isolation Trees, and measure the average path length for each window. Windows considered abnormal are those with the smallest path length. The two main parameters to configure are the number of trees and the proportion of window to draw from the training set used to build each tree. We chose this algorithm for its widespread use, detailed documentation, and minimal parameter tuning.
	No matter the absolute performance of the algorithm, if improvements are observed with executions of the Isolation Forest algorithm with the wavelet-based features compared to executions without these features, it will indicate that incorporating the wavelet transform is useful to detect anomalies in our time series.

	We varied the number of trees between 50 and 1000, and adjusted the proportion of the training set used to build each tree from 25\% to 100\%. Between iterations with 50 trees per forest and iterations with 1000 trees per forest, the mean accuracy score only improved by 0.01. 
	To reduce computation time, we kept the default parameters of Isolation Forest from \textit{scikit-learn}, i.e., 100 trees per forest and 100\% of the training set used for tree building.

	\subsection{Performance of anomaly detection and features' contributions}
	
	The performance of the anomaly detection was assessed by calculating recall, precision, accuracy and F1-score, at each iteration and with or without wavelet-based features. The True Positives are denoted as $TP$ and correspond to the number of abnormal windows detected as abnormal. The False Positives are denoted as $FP$ and correspond to the number of normal windows detected as abnormal. The True Negatives are denoted as $TN$ and correspond to the number of normal windows detected as normal. The False Negatives are denoted as $FN$ and correspond to the number of abnormal windows detected as normal. 

	The recall score corresponds to the rate of anomalies detected on the total number of anomalies: 
	\begin{equation}
	recall = \frac{TP}{FN+TP}
	\end{equation}
The precision score corresponds to the rate of anomaly detections:
	\begin{equation}
	precision = \frac{TP}{FP+TP}
\end{equation}
The accuracy score corresponds to the rate of correct predictions:
	\begin{equation}
	accuracy = \frac{TP+TN}{TP+FN+TN+FP}
\end{equation}
The F1-score is a harmonic mean of the precision and the recall scores:
	\begin{equation}
	F1 = \frac{2TP}{2TP + FP + FN}
\end{equation}
	
We compared the results obtained by running the Isolation Forest algorithm with and without the wavelet-based features to assess the added value of wavelet-based features for anomaly detection. To account for score variability across different splits of the data, 70 separations of training and testing sets were performed. We ran up to 20 iterations of the Isolation Forest for each data split. If the average accuracy over all iterations stabilized before the 20 iterations occurred (i.e., did not change by more than 0.001 in the last five iterations), the loop was terminated early, and the next train/test split continued. The same method was applied for both scenarios: with and without the wavelet-based features.
	
	
We used the SHAP measure to evaluate the features' contribution in the Isolation Forest model (\cite{SHAP}). To account for variability in the SHAP results across different iterations, we conducted 10 separations of the training and testing set. For each separation, we ran 50 iterations of the Isolation Forest. In each iteration, the mean SHAP value was extracted for all the features. At the end of the process, we generated a critical difference diagram (\cite{janez2006}). The diagram illustrates the average rank of each feature's contribution to anomaly detection when applied on a specific data set.
	
	\subsection{Measure of the distance of detection}
\begin{figure}[ht!]
	\begin{subfigure}{1\textwidth}
		\centering
		\includegraphics[width=1\linewidth]{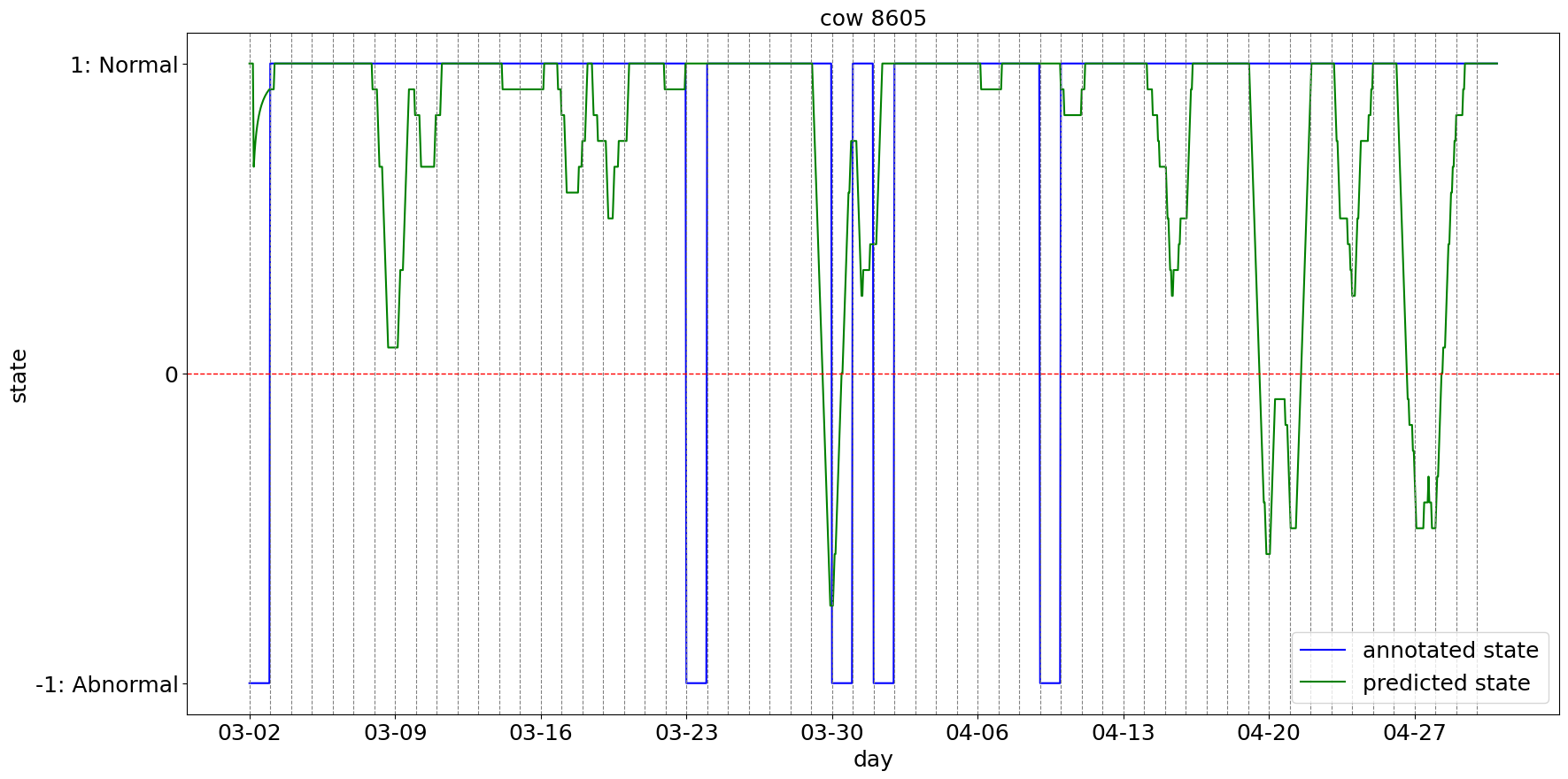}  
		\caption{States of the time series of Cow n°8605. In this time series, some anomalies are annotated by the caretakers.}
		\label{Anomalies_TS1}
	\end{subfigure}
	\begin{subfigure}{1\textwidth}
		\centering
		\includegraphics[width=1\linewidth]{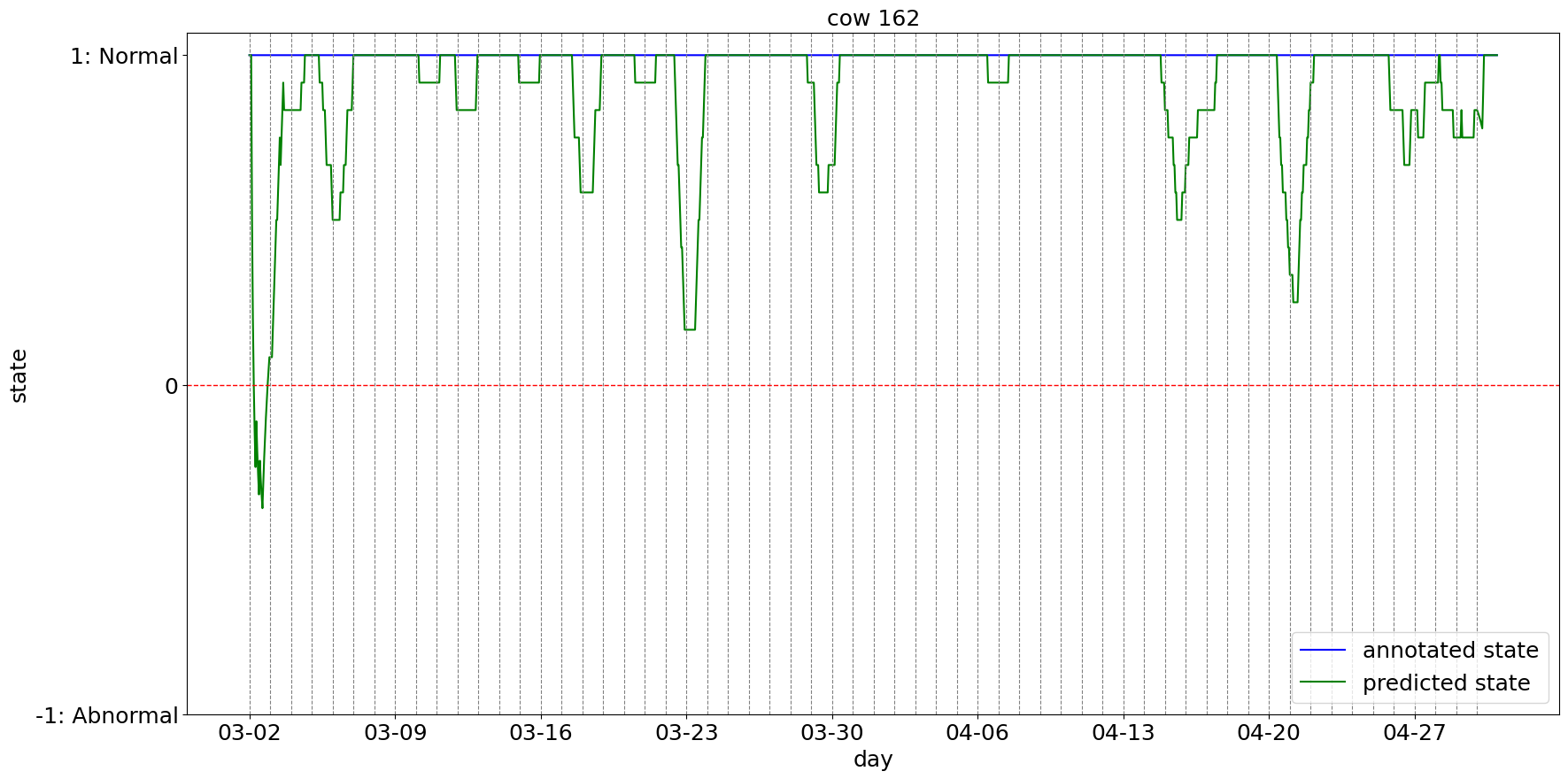}  
		\caption{States of the time series of the Cow n°162 of the Dataset 2. In this time series, no anomaly is annotated by the caretakers.}
		\label{Anomalies_TS2}
	\end{subfigure}
	\caption{States of the time series of two cows from Dataset 2. The time series begin on 2 March 2015 and ends on 30 April 2015. The blue curves represent the states annotated by the caretakers. The green curves represent the predicted states. The red dotted line represents the threshold under which the hours are predicted as abnormal.}
	\label{Anomalies_TS}
\end{figure}

	We set up a new method to identify abnormal days from the abnormal windows to analyze the distance between the detected and annotated anomalies. For each cow, we consider the whole time series of its observation, regardless of the data in the training or testing set. Since each window in the series has a size $q=24$, each hour in the series is contained in at most 24 windows.
	For each hour, we compute the ``predicted state’’ value. It represents the sum of the windows containing this hour that are considered normal minus the sum of the windows containing this hour that are considered abnormal, and the whole divided by the number of windows containing this hour. The predicted state value goes from -1 to +1. We then consider a threshold $\theta \in [-1,1]$. If the predicted state value is below $\theta$, the hour is predicted as abnormal. If at least 12 hours are predicted as abnormal for a certain day, the day is defined as abnormal. Figure \ref{Anomalies_TS} shows the whole time series of a cow with the annotated and the predicted labels for each hour.
	
	The threshold below which each hour is considered as abnormal is $\theta = 0$.
	For each day detected as abnormal, we compute the distance of detection, which represents the numbers of days of offset with the nearest day annotated as abnormal by caretakers. If this distance is negative, it is detected before. If this distance is positive, it is detected after.
	
	\section{Results \& Discussion}
	
	\subsection{Improvement in anomaly detection}

	After computing features for all windows, the correlations between features were checked. If two features were highly correlated (correlation coefficient > 0.9), the feature with the highest average correlation with other features was removed.
	
	Table \ref{results} lists the mean and standard deviation of scores to measure the performance of the detection of anomalies, with and without the wavelet-based features, for the two datasets. 
	
	\begin{table}[ht!]
		\caption{Mean and standard deviation of selected scores of anomaly detection performance, without and with wavelet-based features (WBF).}
		\label{results}
		\centering
		\begin{tabular}{|c|c|c||c|c|}
			\cline{2-5}
			\multicolumn{1}{c}{} &\multicolumn{2}{|c||}{Dataset 1} & \multicolumn{2}{c|}{Dataset 2} \\ \cline{2-5}
			\multicolumn{1}{c|}{}&Without WBF & With WBF & Without WBF & With WBF \\ \hline
			Accuracy&0.51 ± 0.01 & \textbf{0.54} ± 0.01  & 0.54 ± 0.03  & \textbf{0.55} ± 0.03  \\
			Recall& 0.12 ± 0.02 & \textbf{0.13} ± 0.01  & \textbf{0.14} ± 0.01  & 0.12 ± 0.01  \\ 
			Precision& 0.54 ± 0.07 & \textbf{0.67} ± 0.07 & 0.64 ± 0.14 & \textbf{0.68} ± 0.15\\
			F1-score & 0.17 ± 0.02 & \textbf{0.22} ± 0.02 & \textbf{0.22} ± 0.02&0.21 ± 0.02\\
			\hline
		\end{tabular}
	\end{table}

	Without wavelet-based features, the accuracy of the detection is around 0.50 in both datasets, the recall scores are slightly above 0.10, and the precision is around 0.60, resulting in F1-scores around 0.20. These values are definitively low and would not allow a proper detection of cows’ specific states. 
	Adding wavelet-based features results in an improvement of some scores. The accuracy and the F1-score are slightly improved in Dataset 1 (+0.03 and +0.05 i.e. +5.9\% and +29.4\%). The use of wavelet-based features mostly improve the precision in Dataset 1 (+ 0.13 i.e. +24.1\%) but not in Dataset 2. As a consequence, F1-score increased with the use of wavelet-based features only in Dataset 1 (+0.05 i.e. +29.4\%). In general, the use of wavelet-based features improved the performance scores only in Dataset 1. Dataset 2 corresponds to cows subjected to an experimental protocol requiring numerous manipulations that may have hidden behavioral changes due to specific conditions of cows (\cite{lardy_discriminating_2023}).
	
	In Dataset 1, the improvement of the detection thanks to including wavelet-based features is essentially in precision. With a low rate of anomalies, as in the case of specific cow states (prevalence, 0.7\%  in the whole dataset), a high precision is necessary to avoid many false detections which can result in users not paying attention to the anomalies detected by the system. Therefore, wavelet-based features are likely to improve the quality of the detection as perceived by a user of a Precision Livestock system. 
	
	In our study, the testing dataset was constructed to include a balanced number of windows labelled normal and abnormal. This approach gives a fair evaluation of the performance of the anomaly detection algorithm but it does not reflect real-world scenarios where the distribution of normal and abnormal instances is generally skewed. Consequently, the accuracy of our model should not be compared to a random detection of anomalies.  For instance, in Dataset 1, specific states to be detected represented only 2.2 \% of the cow*days. A precision of 0.67, as obtained with wavelet-based features in Dataset 1, would result in the vast majority of alarms received by users as false alarms. However, this is not considering that a detection during fuzzy days can also be valid (see next section).

	\subsection{Distance of detections of anomalies}

\begin{figure}[ht!]
	\begin{subfigure}{.8\textwidth}
		\centering
		\includegraphics[width=.8\linewidth]{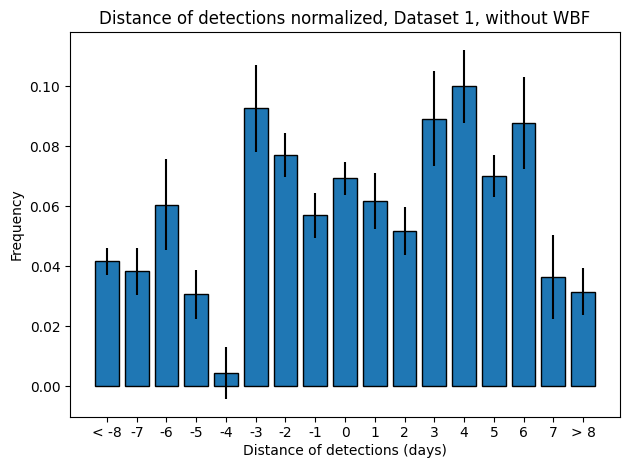}  
		\caption{Without Wavelet-based features (WBF)}
		\label{fig:DistDetectSARA1}
	\end{subfigure}
	\begin{subfigure}{.8\textwidth}
		\centering
		\includegraphics[width=.8\linewidth]{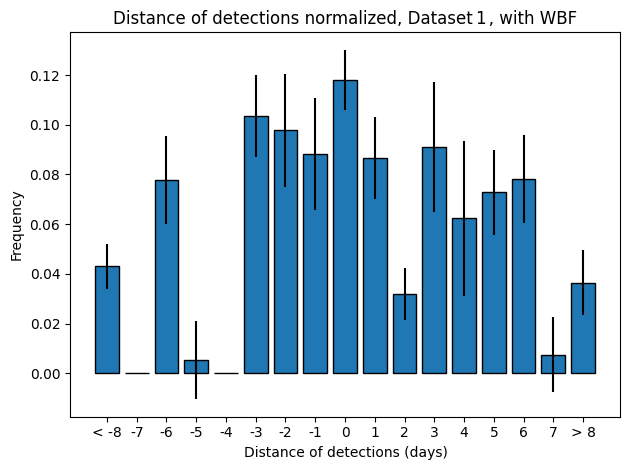}  
		\caption{With Wavelet-based features (WBF)}
		\label{fig:DistDetectSARA2}
	\end{subfigure}
	\caption{Histograms of the mean ratio of the average distance of detection normalized by the initial distribution of the data, for Dataset 1. The black bars represent the standard deviation.}
	\label{fig:DistDetectSARA}
\end{figure}

\begin{figure}[ht!]
	\begin{subfigure}{.8\textwidth}
		\centering
		\includegraphics[width=.8\linewidth]{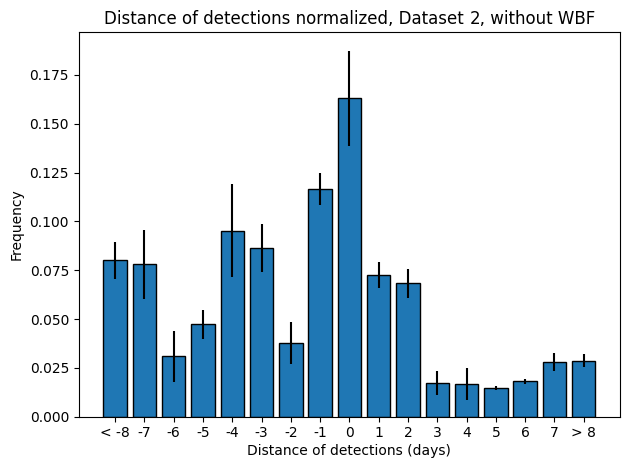}  
		\caption{Without Wavelet-based features (WBF)}
		\label{fig:DistDetectLH1}
	\end{subfigure}
	\begin{subfigure}{.8\textwidth}
		\centering
		\includegraphics[width=.8\linewidth]{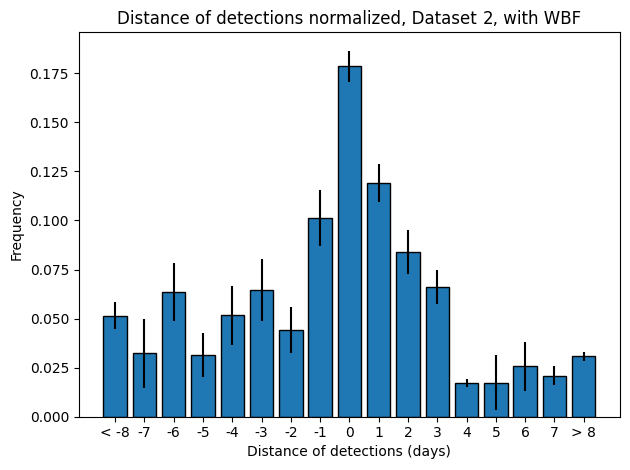}  
		\caption{With Wavelet-based features (WBF)}
		\label{fig:DistDetectLH2}
	\end{subfigure}
	\caption{Histograms of the mean ratio of the average distance of detection normalized by the initial distribution of the data for Dataset 2. The black bars represent the standard deviation.}
	\label{fig:DistDetectLH}
\end{figure}

	The figures \ref{fig:DistDetectSARA} and \ref{fig:DistDetectLH} show the histograms of the distance of detections for the two datasets, with and without considering the wavelet-based features. The histograms have been normalized accounting for initial distribution of the data. The histograms of Dataset 2 form a approximate Gaussian distribution around 0, that is the day when caretakers noted the specific cow state. In Dataset 1, the distance of detection is more variable, nevertheless we notice that the four highest frequencies of detected days are between -3 and +1, when using the wavelet-based features. In Dataset 1, without the wavelet-based features (Figure \ref{fig:DistDetectSARA1}), 35.79\% of the days detected as abnormal have a distance of detection between -3 and +1, and with the wavelet-based features (Figure \ref{fig:DistDetectSARA2}), this rate is 49.40\%. In Dataset 2, without the wavelet-based features (Figure \ref{fig:DistDetectLH1}), 47.65\% of the days detected as abnormal have a distance of detection between -3 and +1, and with the wavelet-based features (Figure \ref{fig:DistDetectLH2}), this rate is 50.77\%. In previous studies (\cite{wagner_detection_2021}), fuzzy days ranged from $\Delta \tau_b = 2$ to $\Delta \tau_a = 7$. Our results suggest that at least one more day should be considered fuzzy, with $\Delta \tau_b = 3$, as the behaviour of the cow may already have changed.

	\subsection{Features contribution}
	
\begin{figure}[ht!]
		\centering
		\includegraphics[width=\textwidth]{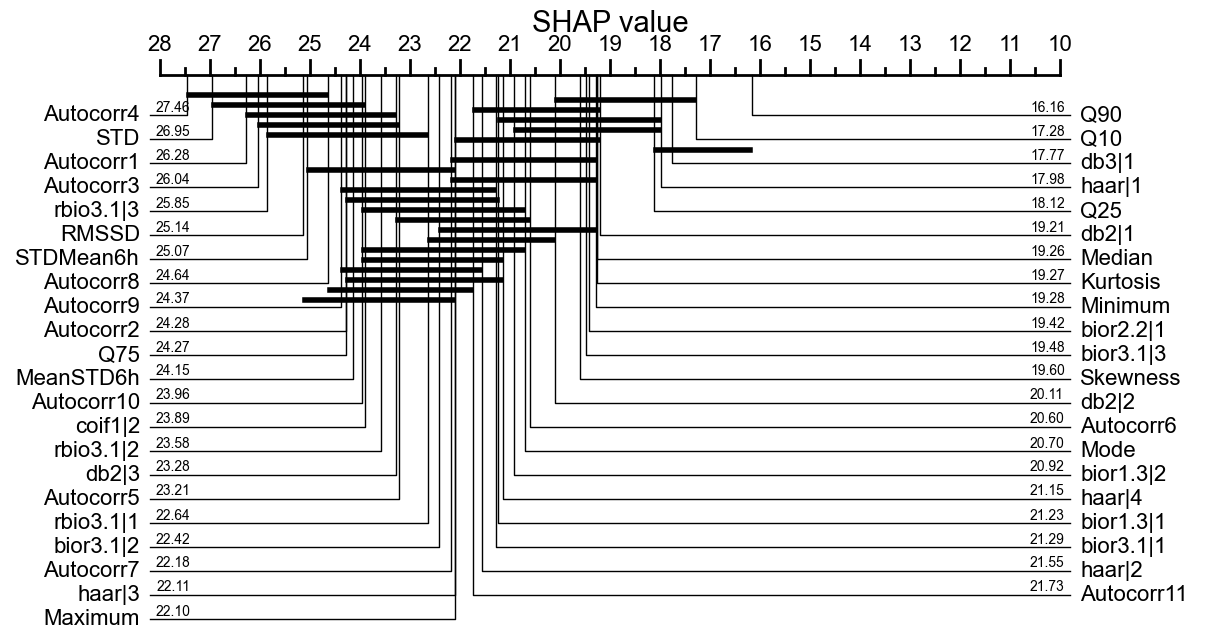}
		\caption{Critical difference diagram for Dataset 1 to measure feature contribution. The value associated with each feature corresponds to its average rank after one execution of the Isolation Forest. The black bars represent features whose contributions are not significantly different. The wavelet-based features are those in the form "\textbf{\textit{waveletname}|\textit{l}}" where \textit{l} is the level of approximation used. These results were obtained with the code in \cite{IsmailFawaz2018deep}.}
		\label{cd-diagram-SARA}
	\end{figure}

In Dataset 1, four of the ten features that contribute the most to the detection are wavelet-based features (Figure \ref{cd-diagram-SARA}). Two of them are among the five top features: Daubechies 3 and Haar, both with one level of approximation.

\begin{figure}[ht!]
		\centering
		\includegraphics[width=1.0\textwidth]{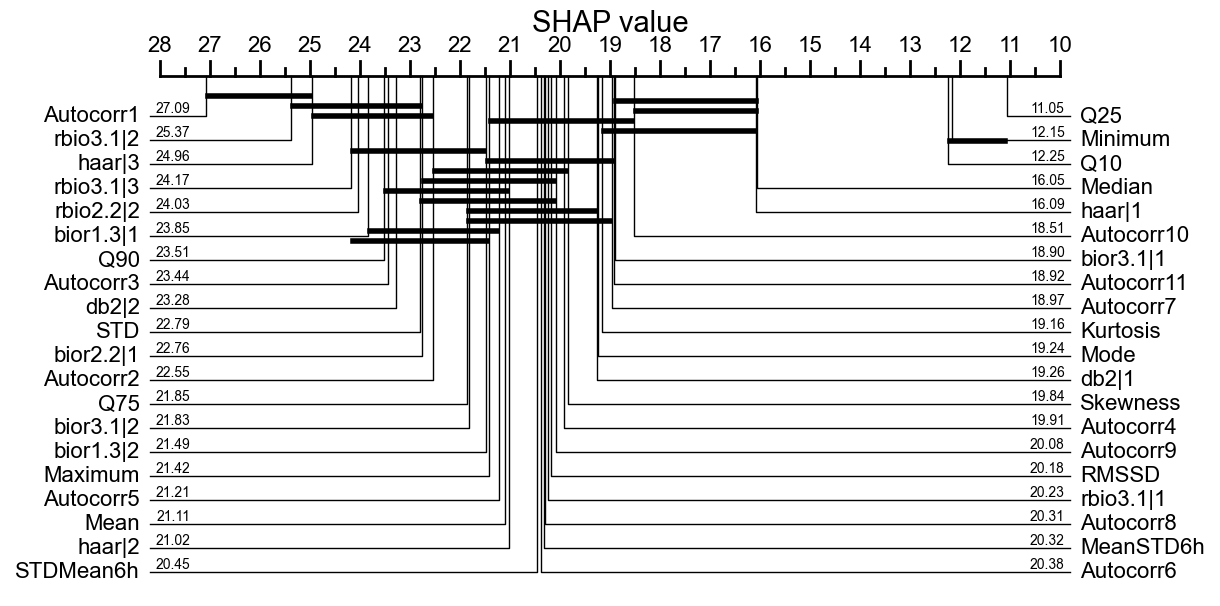}
		\caption{Critical difference diagram for Dataset 2 to measure feature contribution. The value associated with each feature corresponds to its average rank after one execution of the Isolation Forest. The black bars represent features whose contributions are not significantly different. The wavelet-based features are those in the form "\textbf{\textit{waveletname}|\textit{l}}" where \textit{l} is the level of approximation used. These results were obtained with the code in \cite{IsmailFawaz2018deep}.} \label{cd-diagram-LongHealth}
	\end{figure}

In Dataset 2, two of the ten features that contribute the most to the detection are wavelet-based (Figure \ref{cd-diagram-LongHealth}). Haar, with one level of approximation, is one of the five top features. In both data sets, Q25, Minimum, Q10, Median, Kurtosis, and Haar|1 are among the top ten features with the most contributions. Haar|1 makes a significant contribution in both data sets. The wavelet-based features we developed are a good indicator for anomaly detection. Despite being the oldest, the Haar wavelet is the best for these features. In addition, one level of approximation works best, which is the lowest level of approximation of the signal. This means that removing some variations in the time series is relevant to smooth curves while avoiding the loss of too much information.
	
	\section{Conclusions}
	
	Anomaly detection in time series is a well-explored challenge, relevant across various applications and of special interest for Precision Livestock Farming. In many cases, the objective beyond merely detecting anomalies is to understand what makes the anomaly. Transforming time series data into features provides such insights and enhances the explainability of detected anomalies. This paper presents a structured approach to splitting training and testing datasets from time series of cow behavior and proposes the introduction of new features based on wavelet transforms. The features allow to address periodic time series with high variability as they can reduce noise while keeping the main features of a periodic signal. Although the datasets used are complex, we observed that incorporating wavelet-based features slightly increased the detection performance, especially in terms of precision. Thus, the wavelet transform seems promising for the detection of abnormal 24 h cycles of cow's behavior. In addition, many apparent false positive correspond to detection before the anomaly was detected by caretakers, which is of high importance for livestock management.  Detecting an anomaly before caretakers notice allows to take action at an early stage, e.g. taking measures so that a cow health does not deteriorate.
	
	In this study the Isolation Forest was chosen due to its straightforward implementation. More efficient algorithms need to be applied, such as convolutional neural networks (CNNs), including Auto-Encoders (\cite{yao_regularizing_2023}) and Generative Adversarial Networks (GANs) (\cite{jiang_gan-based_2019}). These algorithms will be tested to include the whole procedure of the windows creations, dataset construction, and the wavelet-based features incorporation. Also, our scores of performance focus essentially on instances labeled as normal or abnormal. Fuzzy labels were neglected, except that we considered whether an apparently false detection occurred on a fuzzy day. Performance metrics should be designed to integrate fuzzy labels and a timing factor to provide a more comprehensive assessment of detection quality, in line with Precision Livestock Farming where one wants to warn farmers as early as possible (\cite{wagner_detection_2021}).
	
	\section{Aknowledgements} The authors gratefully acknowledge the support received from the 'Agence Nationale de la Recherche' of the French government through the ‘Investissements d'Avenir’ program (16-IDEX-0001 CAP 20-25). L.E.C.R aknowledges partial financial support from the TOURNESOL programme (FWO/French Embassy in Belgium – Campus France(VS00123N)).
	
	\printbibliography
\end{document}